\documentclass[final,12pt]{clear2026} 

\title[RDMDL]{Bivariate Causal Discovery Using Rate-Distortion MDL: \\ An Information Dimension Approach}
\usepackage{times}
\usepackage{algorithm}
\usepackage{algpseudocode}
\usepackage{multirow}
\usepackage{booktabs}
\usepackage{makecell}
\usepackage{comment}

\clearauthor{%
 \Name{Tiago Brogueira} \Email{tiago.brogueira@tecnico.ulisboa.pt}\\
 \Name{Mário A. T. Figueiredo} \Email{mario.figueiredo@tecnico.ulisboa.pt}\\
 \addr Instituto de Telecomunicações\\
       Instituto Superior Técnico\\
       Universidade de Lisboa, Portugal}

\begin{document}

\maketitle

\begin{abstract}%
Approaches to bivariate causal discovery based on the \textit{minimum description length} (MDL) principle approximate the (uncomputable) Kolmogorov complexity of the models in each causal direction, selecting the one with the lower total complexity. The premise is that nature’s mechanisms are \textit{simpler} in their true causal order. Inherently, the description length (complexity) in each direction includes the description of the cause variable and that of the causal mechanism. In this work, we argue that current state-of-the-art  MDL-based methods do not correctly address the problem of estimating the description length of the cause variable, effectively leaving the decision to the description length of the causal mechanism.  Based on \textit{rate-distortion theory}, we propose a new way to measure the description length of the cause, corresponding to the minimum rate required to achieve a distortion level representative of the underlying distribution. This distortion level is deduced using rules from histogram-based density estimation, while the rate is computed using the related concept of information dimension, based on an asymptotic approximation.
Combining it with a traditional approach for the causal mechanism, we introduce a new bivariate causal discovery method, termed \textit{rate-distortion} MDL (RDMDL). We show experimentally that RDMDL achieves competitive performance on the T\"ubingen dataset. All the code and experiments are publicly available at 
\url{github.com/tiagobrogueira/Causal-Discovery-In-Exchangeable-Data}.
\end{abstract}


\section{Introduction: Bivariate Causal Discovery}
Science seeks to identify cause-and-effect relationships, not just to understand the world, but to intervene on it and answer 'what if' questions \citep{pearl2009causality}. Traditionally, causal relationships are inferred via experiments, either deliberate (e.g., randomized trials) \citep{guo2020survey} or natural \citep{Leatherdale02012019}. When such experiments are infeasible, impractical, unavailable, or unethical \citep{glymour2019review}, as is often the case, \textit{causal discovery} aims at learning causal structure from purely observational data \citep{Eberhardt, icm, zanga2022survey}.

In causal discovery, a key distinction arises between time-series data, where temporal ordering provides causal cues, versus exchangeable data (usually assumed to be independent, identically distributed -- i.i.d.), where no such temporal information exists \citep{timeseriesvsiid, dofinetti}. In the latter, when considering only two variables ($X,Y$), both causal directions ($X \to Y$ and $Y \to X$) belong to the same Markov equivalence class (MEC), which is the smallest causal graph that can be deduced from i.i.d. data with no further assumptions \citep{spirtes2001causation}. 
The bivariate case, also known as the cause-effect problem, has emerged as the quintessential causal discovery problem and has received considerable attention \citep{tuebingenresults}. While general causal discovery seeks to recover entire causal graphs, often validated on synthetic or pseudo-real datasets due to the scarcity of real-world benchmarks, bivariate causal discovery often relies on stricter assumptions on the underlying causal mechanisms, allowing for stronger empirical validation \citep{Eberhardt, ges, tuebingen, tuebingenresults, icm, zanga2022survey}.

A common assumption in bivariate causal discovery (which we adopt) is \textit{causal sufficiency}, \textit{i.e.}, the absence of hidden confounders, thus excluding the possibility of hidden confounders being responsible for the observed dependency between the two variables \citep{tuebingenresults,bayesianmdl}. The most common approaches to tackle the cause-effect problem are score-based, casting the problem as optimizing an objective function. This objective function embodies desired properties of the causal system \citep{zanga2022survey}, and usually is either: a measure of independence between system components, \textit{e.g.}, the cause and the causal mechanism \citep{zhang2010distinguishing,mariouniform}; a measure of fitness to certain model assumptions \citep{cgnn,reci}; a measure of model complexity, \textit{e.g.}, the distance to a non-informative distribution \citep{rkhs} or via the \textit{minimum description length} (MDL) principle \citep{slope,bcqd}).

\section{Background and Contributions}
\subsection{Independent Causal Mechanism, Kolmogorov Complexity, and MDL}
\label{sec:ICM}
The \textit{independent causal mechanism} (ICM) principle states that the causal generative process is composed of autonomous modules that do not inform or influence each other. It is a basic assumption underlying the possibility of causal discovery from observations \citep{Papineau,pearl2009causality,icm}. The ICM principle can be formalized using the notion of (Kolmogorov) \textit{algorithmic independence} as stating that, if random variable $X$ causes $Y$, then the probability distribution of the cause, $P(X)$, and the causal mechanism $P(Y | X)$ are \textit{algorithmically independent} \citep{Janzing2010}, i.e. knowing $P(X)$ does not enable a  \textit{shorter description} of $P(Y | X)$ or vice versa. It has been shown \citep{kolmogorov,Janzing2010} that this algorithmic independence condition implies an inequality between the Kolmogorov complexities (KC) of the two possible causal directions,
\begin{equation} \label{kolmogoroveq}
    K(P(X)) + K (P(Y | X)) \overset{+}{\le} K(P(Y)) + K(P(X | Y)),
\end{equation} 
\noindent 
where $\overset{+}{\le}$ stands for ``less or equal up to a constant". Since KC is uncomputable \citep{LiVitanyi}, the criterion in Eq. \eqref{kolmogoroveq} can not be used directly, and some alternative is needed. For example, \citet{janzing2012information} proposed an information-geometric approach to avoid the uncomputable KC. 

{\noindent\bf Note:} in the next paragraph, we will overload the notation $X$ and $Y$, which will also be used to denote sets of $N$ i.i.d. paired samples from random variables $X$ and $Y$, respectively. That is $X = (x_1,...,x_N)$ and  $Y = (y_1,...,y_N)$.

\citet{kolmtomdl} proposed using the MDL principle as a proxy for the uncomputable KC, making the following two concessions: 1) while KC is not restricted to a specific family of models,   MDL requires a well-defined class of models; 2) whereas KC measures the shortest description of the underlying probability distribution, MDL computes the shortest description of the data (set of samples). They argue that the approximation is valid if the family of models is sufficiently expressive to capture the underlying causal mechanisms and the sampled data is representative of the underlying distribution. The MDL principle for the cause-effect problem corresponds to choosing the $X\rightarrow Y$ direction if $L(X\rightarrow Y) < L(Y\rightarrow X)$, where (and reciprocally for $L(Y\rightarrow X)$),
\begin{equation}
    L(X\rightarrow Y) = L(M) + L(X | M) + L(Y | M,X), \label{eq:LXY}
\end{equation}
where $L(\cdot)$ denotes \textit{description length} (\textit{e.g.}, in bits) and $M$ denotes the model. $L(X | M)$ is the description length of the samples of the cause variable, and $L(Y|X,M)$ is that of the samples of the effect, given the corresponding cause samples. Score-based approaches using the MDL principle include two of the best performing methods: \textit{bivariate quantile causal discovery} (bCQD, \citet{bcqd}) and SLOPE \citep{slope}. 

Recall that MDL was originally proposed as a criterion to select a model in some model class $\mathcal{M}$, according to 
\begin{equation}
\widehat{M} = \arg\min_{M\in\mathcal{M}} L(M) + L(D|M),
\end{equation}
where $D$ is the observed data \citep{originalmdl}. Intuitively, a more complex model (higher $L(M)$) describes the data more precisely (lower $L(D|M)$), while a simpler model (lower $L(M)$) leaves more structure unexplained in the data (higher $L(D|M)$).  MDL elegantly formalizes this trade-off, offering a non-arbitrary criterion based on information theory \citep{originalmdl,Grunwald}.  

\subsection{Computing Conditional Description Lengths}

In this section, we discuss how several MDL-based approaches to bivariate causal discovery define the encoding length of the conditional distribution, i.e. $L(Y|X)=L(Y|X,M)+L(M)$. Since this paper focuses on continuous variables, we only address in detail methods for this type of data. Nevertheless, we mention that \citet{discretecase} deal solely with discrete data and use \textit{stochastic complexity} \citep{Grunwald,Rissanen2007} to define the required description lengths. Stochastic complexity is the negative logarithm of the \textit{normalized maximum likelihood} (NML) distribution, and it is minimax optimal: even if the true distribution is not in the model class, the NML-based stochastic complexity is optimal relative to that class \citep{Grunwald}. 

For continuous variables, to the best of our knowledge, there are four methods that use MDL for bivariate causal discovery: bQCD \citep{bcqd}, COMIC (Bayesian COMpression-based approach to identifying the Causal direction,  by \citet{bayesianmdl}), LCUBE \citep{lcube}, and SLOPE \citep{slope} and variations thereof. 

We start by examining how these methods compute $L(Y | X,M)$. SLOPE uses a combination of global and local regression, both minimizing the sum of squared errors, implicitly corresponding to maximizing the likelihood under Gaussian noise. The residuals are thus encoded under this Gaussian assumption, yielding $L(Y | X,M) \propto \hat{\sigma}^2\left[P(Y | X,M)\right]$, the estimated variance of the conditional distribution $P(Y | X,M)$. Since the objective is to minimize the encoding length, encoding the residuals as a Gaussian distribution is only optimal when these follow a Gaussian distribution \citep{mdlbook}. Similarly, LCUBE also minimizes the sum of squared errors, encodes the residuals under a Gaussian assumption, with the key difference with respect to SLOPE being the choice of cubic regression splines as model functions.

\citet{bcqd} use the \textit{pinball loss} (a.k.a.~\textit{quantile loss}, which is an asymmetric weighted $\ell_1$ loss denoted as $S$)) to fit the model and encode the residuals using the asymmetric Laplace distribution,
\begin{equation}\label{bqcd_eq}
    L(Y | X,M) \propto \sum_{j=1}^m w_{\tau_j} \sum_i^N S(Y_i,q_{X,\tau_j})
\end{equation}
\noindent where $\tau_j$ indicates the j-th quantile, $q_{X,\tau_j}$ indicates the value of the estimated j-th quantile of Y based on $(X,\tau_j)$, and m is the total number of quantiles in consideration. 

Finally, COMIC \citep{bayesianmdl} uses a Bayesian formulation to express the model, 
\begin{equation}
    L(Y | X,M) =-\log \int p(Y | X, \theta) p(\theta) d\theta ,
\end{equation}
\noindent where $p(\theta)$ is estimated using mean-field variational inference.

Regarding $L(M)$, COMIC and bQCD both abstain from computing it. In COMIC, $p(\theta)$ (the model) is computed from the data, under a Gaussian approximation, so there is no model to encode. For bQCD, they argue that both causal directions require encoding the same number of quantiles, so it is a constant that can be ignored. 

LCUBE assumes parameter independence, adopting the classic choice \citep{mdlbook}
\begin{equation}\label{paramslength}
    L(M) = L(\theta) = (k/2) \log_2(N),
\end{equation}
where $\theta$ is the $k$-dimensional vector of model parameters and $N$ the number of samples.

Lastly, SLOPE computes the size of the model description by adding the combinatorial encoding representing the choice of models, both for the global and each of the local regressions, with a fixed \textit{a priori} precision of each parameter within the model.


\subsection{Analysis of $L(X)$}
\label{sec:critiqueLx}
This subsection describes how each state-of-the-art MDL method computes $L(X)$, pointing out their restrictive assumptions. As described in Subsection \ref{sec:ICM}, for MDL to approximate well the Kolmogorov complexity, the chosen family of models must be representative of the true distribution. As we argue next, bQCD, COMIC, LCUBE, and SLOPE do not do so when defining $L(X)$. 

LCUBE assumes a uniform prior over a normalized distribution (both for $X$ and $Y$), thus explicitly setting $L(X)=L(Y)$. SLOPE assumes a discretized uniform distribution over the minimum required resolution, thus ignoring any relevant information about the distribution. COMIC measures $L(X)$ by encoding $X$ as the residuals of a Gaussian distribution with the mean estimated from the samples; as SLOPE and LCUBE, this ignores any information regarding the shape of the distribution. Finally, bQCD computes multiple estimates at its quantiles, thus assuming a necessarily discrete underlying distribution. 

GPI \citep{kolmogorov} also seeks to approximate Kolmogorov complexity, but using Bayesian theory instead of MDL. Both approaches balance the complexity of the cause and conditional distributions. Thus, an analogy can be made where the negative log-likelihood of the cause distribution in Bayesian theory corresponds to MDL's cause distribution encoding length (and similarly for the conditional distribution). For the latter, the conditional log-likelihood in GPI takes up a complex formula that we decided not to dive into. However, given this work's emphasis on $L(X)$, it is important to mention how GPI computes the log-likelihood of the observed cause distribution. First, it estimates a Gaussian mixture model from the data, and then measures the log-likelihood of the observed points according to the minimum message length principle for Gaussian mixtures defined by \citet{mml}. Consequently, it measures the distance of the observed distribution to the best-fitted Gaussian mixture. Intuitively, it states that the more similar the observed distribution is to a Gaussian mixture, the smaller its complexity. Additionally, throughout this paper, even though GPI is not an MDL-based method, we will often refer to it as such, alongside the other four analysed (bQCD, COMIC, LCUBE, and SLOPE). This will be useful, since GPI shares MDL's rationale of measuring the complexity of both the cause and the causal mechanism. Furthermore, for GPI, $L(X)$ can be defined as $-\log p(X)$.

In all, none of these methods attempt to capture features of $P(X)$, or to approximate in a meaningful way the real distribution. Instead, they assume $X$ follows a concrete distribution (e.g., Uniform, Gaussian, discrete 
Uniform, or mixture of Gaussians\footnote{Although a Gaussian mixture  is theoretically capable of approximating any probability distribution arbitrarily well given infinitely many observations, in practice, since each example contains few datapoints, the approximation only holds when the true underlying distribution is itself a mixture of Gaussians.}.), resulting in poor approximations of the Kolmogorov complexity of $P(X)$, when these narrow assumptions do not hold.

More generally, note that the use of MDL (or Bayesian theory) in the problem of causal discovery (see Equation \eqref{eq:LXY}) is unorthodox, as the need to estimate and compare the complexity of the cause distribution is uncommon. For example, consider the use of MDL to perform model selection in a regression problem; \textit{e.g.}, selecting the order when fitting a polynomial to a collection of $(x,y)$ pairs, simply written as $(X,Y)$. In this case, the MDL criterion would read
\begin{equation}\label{mdl_standard}
    \widehat{M} = \arg\min_{M\in\mathcal{M}} L(M) + L(X) + L(Y|X,M)   = \arg\min_{M\in\mathcal{M}} L(M) + L(Y|X,M),
\end{equation}
where, here, $M$ specifies only the regression model (\textit{e.g.}, the polynomial order), thus $L(X)$ does not depend on $M$. The second equality, i.e., dropping $L(X)$, is justified by the fact that, in this case, it is a constant, since the description length of the independent variable does not depend on the regression model. In contrast, in the bivariate causal discovery problem, we are comparing two models in which $X$ and $Y$ play different roles; consequently, when using MDL to select between the two possible causal directions, the description lengths of the cause $L(X)$ (in the $X\rightarrow Y$ model) and $L(Y)$ (in the $Y\rightarrow X$ model) do not cancel each other out.

Although this difference between bivariate causal discovery and more standard applications of MDL  partly explains why no method uses an informative estimate of $L(X)$, it still means that none are faithfully following the MDL framework to the causal discovery problem. By relying almost only on $L(Y | X ,M)$, these methods are close to score-based methods that simply compute the causal direction that best fits a specific regression model (e.g., RECI, \textit{regression error based causal inference}, proposed by \citet{reci}). The difference then lies in MDL-based approaches measuring goodness of fit using an MDL formalism, similarly to the regression problem defined in Equation \eqref{mdl_standard}.  In turn, this deviates from the original motivation behind MDL for causal discovery: approximating Kolmogorov complexity in the two directions (as explained in Subsection \ref{sec:ICM}). 

Dropping the dependency on the model assumptions $M$ for notational simplicity, MDL in causal discovery compares $L(X)+ L(Y | X)$ with $L(Y)+ L(X | Y)$, choosing the direction that corresponds to a smaller encoding. This can be written as computing a score,
\begin{equation} \label{eq:difs}
    s = (L(X) - L(Y) ) + ( L(Y | X) - (L(X | Y) ),
\end{equation}
and choosing $X\rightarrow Y$ if it is $s<0$, and $Y\rightarrow X$ otherwise. Thus, MDL can be seen as the sum of two differences, one representing the difference in encoding length between the cause distribution and another representing the difference in the encoding lengths of the two conditional distributions. In turn, each of these two differences can be seen as an individual score, each attempting to estimate the true causal direction, and MDL resulting in the combination of these two estimators. 

Taking a step back, and noting that the use of MDL in causal discovery stems from the ICM principle, the lengths in the true causal direction will not encode overlapping information over the joint distribution. In contrast, the same cannot be said in the wrong causal direction. More relevant to this work, the length of the effect ($Y$) is expected to encode information over the conditional distribution of $X|Y$, while the opposite is not true. Therefore, even though MDL succeeds due to the complementary nature of these two estimators, each alone should be expected to perform significantly above average. Given that observation, the corresponding $(L(X)-L(Y))$ estimators of four out of the five analysed causal discovery methods were evaluated, using the \textit{area under the receiver operating characteristic} (AUROC), on the T\"ubingen benchmark \citep{tuebingen} (LCUBE was excluded since it defines $L(X)=L(Y)$). Table~\ref{lxperformance} shows that, for all methods, $(L(X) - L(Y))$ by itself has significantly lower AUROC than the corresponding full methods. In turn, this further confirms the notion that current MDL methods for causal discovery fail to make accurate estimates over the encoding length of the cause distribution.

\begin{table}[hbt]
\centering
\caption{AUROC (in \%) on the T\"ubingen dataset, achieved by $(L(X)-L(Y))$ alone as a prediction score versus the full method, for bQCD, COMIC, GPI, and SLOPE.}\label{lxperformance}
{\small \begin{tabular}{lccc}
\toprule
&  $L(X)-L(Y)$  &  full method\\
\midrule
$\text{bQCD}$& 54.7 & 71.1 \\
$\text{COMIC}$  & 55.6 & 76.0\\
GPI & 49.0 & 68.1 \\
$\text{SLOPE}$   & 53.7 & 80.5\\
\bottomrule
\end{tabular}}
\end{table}


\subsection{Contributions and Overview}
In this work, we started by arguing (Subsection \ref{sec:critiqueLx}) that the five previously mentioned state-of-the-art MDL-based methods for the cause-effect problem with continuous variables (bQCD, COMIC, GPI, LCUBE, SLOPE) do not adequately define $L(X)$, thus not following the MDL recipe for causal discovery. In the next section, we propose a new approach based on rate-distortion theory to define $L(X)$ and couple it with a standard approach to compute $L(Y | X)$, yielding the new causal discovery method RDMDL. Subsequently, we overview the main assumptions underlying our approach and report experimental results on the T\"ubingen and on synthetic benchmarks, showing that RDMDL performs competitively with the best methods.

\section{Proposed Approach: RDMDL}
This section describes the proposed MDL-based approach for bivariate causal discovery. The key contribution is a new way to define the description length of the cause, $L(X)$, based on rate-distortion theory.  We then couple it with a standard choice to compute $L(Y | X,M)$, leading to the proposed RDMDL. A detailed description of the algorithm is present in Appendix \ref{algorithm_appendix}. We begin by reviewing the concepts from rate-distortion theory that are needed to describe RDMDL. 

\subsection{The Rate-distortion Function and Information Dimension}
The \textit{rate-distortion} (R-D) function is central in the theory of lossy compression \citep{Cover,ratedistortionintro}. For a random variable $X \in \mathbb{R}$ and using mean squared error (MSE) as the distortion measure, the R-D function is defined as 
\begin{equation} \label{ratedistortioneq}
R(X, D) \triangleq \inf_{P_{\hat{X}|X}: \; E[(X-\hat{X})^2]\le D} I(X ; \hat{X}),
\end{equation}
where $I(X; \hat{X}) = H(X) - H(X|\hat{X}) = H(\hat{X}) - H(\hat{X}|X)$ is the mutual information between $X$ and a reconstruction/approximation $\hat{X}$, $H(X)$ and $H(X|\hat{X})$ denote Shannon differential entropy and conditional entropy, and $D$ is an upper bound on the allowed distortion. Essentially, the R-D function is the minimum amount of information (in bits, if mutual information is computed with base-2 logarithms) that any $\hat{X}$ needs to have about $X$ to satisfy the distortion upper bound. 

The \textit{rate-distortion dimension} (RDD) of a source is defined as  
\begin{equation} \label{rddeq}
\dim_{R}(X) = 2 \lim_{D \downarrow 0} \frac{R(X,D)}{-\log D}
\end{equation}
and measures how the number of bits required to encode a vector scales with the maximum allowed distortion, as that allowed distortion approaches zero \citep{ratedistortiondimension}. For example, any source with an absolutely continuous distribution has $\dim_{R}(X) = 1$, while any discrete source has $\dim_{R}(X) = 0$. Intuitively, for a discrete source, even with a countably infinite alphabet, each outcome lies in an isolated point of the space, so achieving small MSE does not require specifying any continuous degree of freedom. In contrast, a source taking values in a continuum requires resolving a continuous dimension, and describing it to vanishing distortion is reflected in its RDD.

\citet{informationdimensionratedistortionoriginal} proved that the RDD coincides with the so-called \textit{information dimension} (ID, \citet{renyi}), 
\begin{equation}\label{informationdimension_eq}
 \dim_{R}(X) =   \dim_I(X)
= \lim_{\varepsilon \to 0}
H_{\varepsilon}(X)/\log(1/ \varepsilon),
\end{equation}
where $H_{\varepsilon}(X)$ is the (discrete) entropy of a uniformly quantized version of $X$ with resolution $\varepsilon$.

\subsection{Setting $L(X)$} \label{lxsubsection}
We propose to approximate $L(X)$ by $R(X,D)$, for a distortion level $D$ that is representative of the relevant sampled information from the underlying probability distribution. To do so, two questions must be answered: 1) what does it mean for a distortion to be representative of the underlying probability distribution? 2) How does the theoretical lower bound provided by the R-D combines with the MDL approach to approximate the KC of the underlying distribution?

Starting with the second question, MDL approximates the KC by adopting a model, encoding the residuals, and measuring the description length of $X$ as that of the residuals plus the model. Therefore, it approximates the theoretical lower bound of KC by measuring the shortest description length for a specific set of models. As mentioned in Subsection \ref{sec:ICM}, this requires the following two assumptions: the samples are representative of the underlying probability distribution; and the set of models approximates well the real distribution. By using R-D theory to measure $L(X)$, the second assumption is no longer required, as we directly measure a non-parametric lower bound estimate of the encoding length of $X$.  

The other major question is how to choose the distortion level so that $\hat{X}$ is representative of the underlying probability distribution. As is well known,  $\lim_{D \to 0} R(X,D) = \infty$, since we are dealing with continuous variables. Conversely, under MSE distortion and assuming finite variance, $\mbox{var}(X) < \infty$, then $D \geq \mbox{var}(X) \Rightarrow R(X,D) = 0$, which corresponds to setting $\hat{X} = \mathbb{E}(X)$ \citep{Cover}. Between these two extremes, there is a middle ground, where $D$ is small enough such that $\hat{X}$ ``accurately" represents $X$ (and consequently $R(X,D)$ is a good approximation of $L(X)$), while not being so small such that it forces $\hat{X}$ to encode all the randomness inherent to the sampling process. This question is akin to that of optimal bin-size selection for histogram-based non-parametric density estimation \citep{Scott1992}. If the quantization bin width $\delta$ is too large (distortion is large), the histogram has high bias, ignoring genuine structural information about the distribution. Conversely, if $\delta$ (and $D$) is too small, the histogram has high variance, and we are no longer just encoding the density, but also irrelevant fluctuation of individual samples, thus effectively ``wasting bits" to store noise rather than structure.

For histogram-based density estimation, there are several well-established rules for optimal bin sizes. We will consider four of these rules:
\begin{description}\label{possible distortions}
    \item[Freedman-Diaconis' rule:]  $\delta = 2 \; \mbox{IQR}(X)\; N^{-1/3}$, where $\mbox{IQR}(X) = Q3 - Q1$ is the interquartile range of the samples in $X$, \textit{i.e.}, the difference between the 75th percentile and the 25th percentile  \citep{FDRule}.
    \item[Sturges' rule:] $\delta = \mbox{range}(X)/(1+\log_2 (N))$, where $\mbox{range}(X)$ is the range of the samples in $X$ \citep{Sturges,ScottSturges}.
    \item[Scott's rule:] $\delta \simeq  3.5\; \hat{\sigma}(X)\; N^{-1/3}$, where $\hat{\sigma}(X)$ is the sample standard deviation \citep{ScottRule}.
    \item[Rice University rule:] $\delta = 0.5\; \mbox{range}(X)\; N^{-1/3}$ \citep{RiceRule}.
\end{description}

Since the Freedman-Diaconis' rule is the only non-parametric one, it stands out as the most fitting to our approach, as thus, it will be considered as the default implementation of RDMDL.
Howerver, unlike in histogram-based density estimation, here the goal is not to approximate the probability density function, but the samples themselves. Consequently,  using the well-known high-resolution approximation, the MSE resulting for quantizing with bin size $\delta$ is simply $D = \delta^2 /12$, which we will adopt, where $\delta$ is set using one of the four rules just mentioned.

Given a choice of distortion level $D$, we compute $L(X)$ from $R(X,D)$. To do so, we first estimate the ID, which is known to be equivalent to the RDD. Then, from the desired distortion and the RDD, we estimate the rate required to encode $X$. We assume that we are in the high-resolution (small $\epsilon$ and small $D$) asymptotic regime, thus the leading terms of both $R(X,D)$ and $H_\epsilon(X)$ are dominant. Under this assumption, Equations \eqref{rddeq} and \eqref{informationdimension_eq} can be transformed into, respectively, 
\begin{align}
    H_\epsilon(X) &= -\,\dim_I(X) \, \log(\epsilon) \;+\; o(\log(1/\epsilon)), \\
    R(X,D)       &= -\,\frac{\dim_R(X)}{2} \, \log(D) \;+\; o(\log(1/D)),\label{asymptoticapprox}
\end{align}
where $o(\log(1/\epsilon))$ and $o(\log(1/D))$ are stricter upper bounds of any unknown additive terms. 

To estimate the ID, we begin by considering a set of bin sizes \(\{\epsilon_m\}_{m=1}^M\) (small enough to be in the high-resolution regime) and compute the entropy at each scale,
\begin{equation}
H_{\epsilon_m}(X) = - \sum_{j=1}^{N_m} p_j^{(\epsilon_m)} \log p_j^{(\epsilon_m)},
\end{equation}
where \(p_j^{(\epsilon_m)}\) is the probability of \(X\) falling into the \(j\)-th bin of size \(\epsilon_m\), and \(N_m\) is the number of bins at that scale (e.g., if $X\in [0,\, 1]$, then $N_m = 1/\epsilon_m$). The information dimension is then obtained by performing a least-squares fit of \(H_{\epsilon_m}(X)\) against \(\log (1/\epsilon_m)\):
\begin{equation}\label{lseq}
\mbox{dim}_I(X) = \underset{d}{\mathrm{argmin}} \; \sum_{m=1}^M \bigg( H_{\epsilon_m}(X) - d \, \log \frac{1}{\epsilon_m} \bigg)^2.
\end{equation}
Afterwards, since $\mbox{dim}_R = \mbox{dim}_I$ \citep{informationdimensionratedistortionoriginal}, it is possible to compute $R(X,D)$ for the chosen $D$ directly from Equation \eqref{asymptoticapprox}. Finally, $L(X)$ is set to the length needed to encode the $N$ samples in $X$, thus
\begin{equation}\label{lx_calc}
    L(X) = N\, \cdot R(X,D) = N\; \frac{\mbox{dim}_I(X)}{2} \log\!\left(\frac{1}{D}\right).
\end{equation}

\subsection{Setting $L(Y| X)$ and $L(M)$}
The central goal of this paper is to highlight and motivate the focus on $L(X)$ when using MDL for causal discovery. Consequently, a very standard approach is used for $ L(Y | X,M)$ and $ L(M).$ Following the standard assumptions in MDL, and specifically those used in SLOPE and LCUBE, we used least squares to fit the model, and Gaussian encodings to represent the residuals, ensuring coherence between both operations. This choice yields
\begin{equation}
    L(Y | X,M) = \frac{N}{2}\log_2\!\big(2\pi\,\hat{\sigma}^2\big) + \frac{N}{2\log 2},
\end{equation}
where $\hat{\sigma}^2$ is the empirical residual variance. This term corresponds to the negative log-likelihood (in bits) under a Gaussian model with variance $\hat{\sigma}^2$. 

As model, we use a single global regressor from the following three parametric families:
\begin{itemize}\itemsep0.1cm
    \item \textit{Polynomial models}, with degree up to 5.
    \item \textit{Reciprocal models:} $\hat{y} = a/x^p + b$, with $p \in \{1,2\}$.
    \item \textit{Exponential and logarithmic models:} $\hat{y} = a\,e^{x} + b$ and $\hat{y} = a\,\log(x) + b$.
\end{itemize}
The model description length $L(M)$ is set to the standard choice in Equation \eqref{paramslength}, where $k$ is the number of parameters. This choice minimizes the total conditional MDL ($L(Y|X)$), if the model parameters are independent. Finally, for each family of models, all admissible parameterizations are fitted, and their total code length $L(Y | X,M)+L(M)$ is evaluated, and
\begin{equation}
    L(Y | X) = \min_M L(Y | X,M)+L(M).
\end{equation}

\subsection{Overview of RDMDL's assumptions}
All bivariate causal discovery methods share the same goal. What differentiates them is the underlying assumptions of each method. For greater clarity and to facilitate future research, we consider that our method is supported by the following non-trivial assumptions/claims:
\begin{enumerate}
    \item There is an interval of distortions $D$ such that $R(X,D)$ is a good approximation of $L(X)/N$ and histogram-based density estimation rules can compute distortions $D$ that fall within this interval. This claim is supported by the good performance of all RDMDL's variations.
    \item The samples are representative of the underlying probability distributions. It corresponds to the first assumption in Subsection \ref{sec:ICM}, and it is necessary to ensure the equivalence between the description length of the samples and that of the underlying joint distribution.
    \item The family of models used to encode $L(Y | X)$ is representative enough to capture the underlying causal mechanisms. This is the second criterion in Subsection \ref{sec:ICM}. 
\end{enumerate}

\section{Results}
In this section, we start by analyzing how RDMDL compares with other bivariate causal discovery methods and, afterwards, we will study how our approach to computing $L(X)$ contributes to the performance of RDMDL. As described in Section \ref{possible distortions}, we consider four different rules to estimate the optimal bin size (and with it the desired distortion). Consequently, we report the results obtained by each implementation, named RDMDL-FD (using the Freedman-Diaconis rule), RDMDL-R (Rice University rule), RDMDL-Sc (Scott's rule), and RDMDL-St (Sturges rule). Additionally, remember that the FD rule is considered the default implementation of RDMDL, given its non-parametric approach.

Table \ref{lisbonandtuebingenres} shows the results on the T\"ubingen dataset of \citet{tuebingen}, the only real-world bivariate causal discovery benchmark. In turn, method performance was measured using three evaluation metrics: AUROC, accuracy, and AUDRC (area under the decision rate curve \citep{audrc}).
The table includes results for a comprehensive set of other open-source bivariate causal discovery methods. In fact, to the best of our knowledge, all causal discovery methods with open-source and retrievable implementations, in Python or R, are included: ANM \citep{peters2010identifying}, bQCD, CAM \citep{cam}, CDCI \citep{cdci}, CDS \citep{cds}, CGNN \citep{cgnn}, FOM \citep{fom}, GPI, HECI \citep{heci}, IGCI \citep{mian2023information}, LCUBE, LOCI \citep{audrc}, NNCL \citep{nncl}, RECI \citep{reci}, ROCHE \citep{roche}, SLOPE, and SLOPPY \citep{sloppy}\footnote{For bQCD, CAM, CDCI, FOM, GPI, HECI, LCUBE, LOCI, NNCL, ROCHE, SLOPE, and SLOPPY, the original implementations provided by the authors were used. For ANM, CDS, CGNN, IGCI, and RECI, the code from the CausalDiscoveryToolbox of \citet{cdtoolbox} was used. Finally, since the original implementation of CGNN was too slow, the hyperparameters were adapted to ensure no example took longer than $30$ seconds to score.}. For methods without publicly available implementations, the reported results are present in Appendix \ref{unimplementedres}. The only method with an open source implementation that we excluded is GPLVM \citep{bayes}, because it involves training on the same dataset on which it is evaluated. Therefore, it is not a pure causal discovery method, as the model is able to learn the causal relations present in the dataset under evaluation. 

The results in Table \ref{lisbonandtuebingenres} show that all versions of RDMDL outperform most methods on the T\"ubingen dataset across all metrics. The only exceptions are SLOPE, SLOPPY, and RECI, which are still outperformed by RDMDL-FD in terms of AUROC and AUDRC. Moreover, comparing to other results reported in the literature without publicly available implementations, only GRCI, COMIC, and GPLVM outperform RDMDL-R, RDMDL-Sc, and RDMDL-St (while scoring worse than RDMDL-FD). However, similarly to GPLVM, both COMIC and GRCI also require training on the benchmark dataset, and so, their results should be taken with a grain of salt. 

Additionally, the T\"ubingen dataset is known for having more discrete causes than effects. Controlling for such, and scoring only the examples whose number of unique values per variable is at least a third of the total number of points, RDMDL-FD still obtains an AUROC of 84.5\% and the $(L(X)-L(Y))$ estimator alone an AUROC of $73\%$. This accounts for a total weight of $39\%$ of the original dataset, without a significant dip in RDMDL's performance. Therefore, the proposed measure of $L(X)$ captures a notion of complexity of $X$ that goes beyond a mere notion of how ``discrete" it is. Instead, it appears to follow the initial intuition of measuring complexity by assessing how rugged the distribution becomes when using smaller and smaller bins (see subsection \ref{lxsubsection}).

\begin{table}[h]
\caption{Performance of the 4 RDMDL versions   and 17 other methods on the T\"ubingen benchmarks. The best results in each column are in bold; the second best is underlined.}
\label{lisbonandtuebingenres}
\centering
\resizebox{0.5\textwidth}{!}{
\begin{tabular}{l|lll}
\toprule
Method & AUROC (\%) & Accuracy (\%) & AUDRC (\%) \\
\midrule
ANM & 61.9 & 60.4 & 62.9 \\
bQCD & 71.1 & 69.6 & 70.1 \\
CAM & 46.6 & 52.3 & 43.1 \\
CDCI & 60.7 & 61.5 & 55.0 \\
CDS & 59.0 & 60.5 & 57.8 \\
CGNN & 61.6 & 61.5 & 69.6 \\
FOM & 45.1 & 45.5 & 40.9 \\
GPI & 68.1 & 68.7 & 67.0 \\
HECI & 75.3 & 70.5 & 79.2 \\
IGCI & 69.8 & 60.9 & 73.3 \\
LCUBE & 58.0 & 58.9 & 70.1 \\
LOCI & 57.4 & 61.5 & 49.5 \\
NNCL & 65.3 & 55.2 & 63.4 \\
RECI & 76.4 & 70.5 & 75.7 \\
ROCHE & 56.7 & 53.0 & 53.7 \\
SLOPE & \underline{80.5} & \textbf{73.3} & \underline{86.4} \\
SLOPPY & 79.3 & \underline{72.6} & 85.3 \\
\hline
RDMDL-FD (ours) & \textbf{82.0} & 71.9 & \textbf{86.6} \\
RDMDL-R (ours) & 74.7 & 68.0 & 79.1 \\
RDMDL-Sc (ours) & 76.3 & 67.4 & 81.7\\
RDMDL-St (ours) & 75.3 & 68.0 & 80.1 \\
\bottomrule
\end{tabular}
}
\end{table}

Table \ref{syntheticrdmdl} shows the results of RDMDL, bQCD, GPI, LCUBE, and SLOPE on three synthetic benchmarks: CE \citep{tuebingenresults}, ANLSMN \citep{bcqd}, and SIM \citep{tuebingen}. The performance of all other implemented methods is in Appendix \ref{syntheticappendix}. It is clear that all versions of RDMDL perform significantly above average in most synthetic datasets, confirming it is a strong method. However, it is also notable that RDMDL does not achieve extremely high performance in these datasets, unlike bQCD or LCUBE. Usually, causal discovery methods are designed with certain assumptions (\textit{e.g.}, additive noise or uniform distributions), which can more easily be matched with the generation process of synthetic datasets. In contrast, RDMDL is more agnostic; while it achieves very good results on the real-world benchmark, it does not seem to take as much advantage of explicit causal assumptions present in synthetic datasets. 

\vspace{-0.4cm}

\begin{table}[h]
\caption{Results of the four versions of RDMDL in three collections of synthetic datasets: CE, ANLSMN, and SIM, evaluated using AUROC and accuracy (in parentheses in each cell).The best results for each dataset are in bold and the second-best are underlined.}
\label{syntheticrdmdl}
\centering
\resizebox{\textwidth}{!}{
\begin{tabular}{lccccccccccccc}
\toprule
Method &
CE\text{-}Cha & CE\text{-}Gauss & CE\text{-}Multi & CE\text{-}Net &
AN & AN\text{-}s &
LS & LS\text{-}s &
MN\text{-}U &
SIM & SIM\text{-}c & SIM\text{-}G & SIM\text{-}ln \\
\midrule
RDMDL-FD & 62.7 (\underline{60.7}) & 64.7 (58.7) & \textbf{97.0} (\textbf{89.3}) & 75.2 (68.3) & 51.2 (49.0) & 43.3 (45.0) & 49.0 (47.0) & 13.4 (17.0) & 0.9 (6.0) & 53.6 (47.0) & 59.6 (57.0) & 59.4 (55.0) & 87.2 (77.0) \\
RDMDL-R & \underline{64.2} (60.3) & 64.5 (59.0) & 96.5 (88.3) & 81.1 (72.0) & 84.5 (74.0) & 52.3 (51.0) & 77.0 (68.0) & 6.9 (13.0) & 1.3 (7.0) & 61.9 (56.0) & 66.9 (65.0) & 66.2 (57.0) & 91.4 (79.0) \\
RDMDL-Sc & 63.7 (59.3) & 69.4 (62.7) & \underline{96.9} (\underline{89.0}) & 77.5 (69.3) & 58.6 (52.0) & 34.7 (34.0) & 53.5 (51.0) & 4.8 (12.0) & 0.8 (4.0) & 58.1 (53.0) & 63.9 (62.0) & 60.1 (54.0) & 88.6 (78.0) \\
RDMDL-St & 64.0 (60.3) & 65.1 (59.0) & 96.5 (88.3) & 81.7 (72.3) & 86.8 (76.0) & 53.7 (51.0) & 78.5 (69.0) & 5.8 (13.0) & 1.4 (7.0) & 62.8 (56.0) & 67.6 (66.0) & 66.4 (58.0) & 91.6 (79.0) \\
\hline
bQCD & 56.3 (55.0) & 48.1 (54.3) & 53.2 (49.0) & \textbf{90.7} (\underline{81.3}) & \textbf{100.0} (\textbf{100.0}) & \underline{90.5} (\underline{83.0}) & \textbf{100.0} (\textbf{100.0}) & \textbf{100.0} (\textbf{100.0}) & \textbf{100.0} (\textbf{100.0}) & 69.9 (\underline{68.0}) & \underline{80.8} (\underline{77.0}) & 71.7 (67.0) & 91.9 (\textbf{87.0}) \\
GPI & \textbf{64.9} (\textbf{62.4}) & \textbf{89.1} (\textbf{83.2}) & 72.2 (71.5) & \underline{88.8} (\underline{81.3}) & 79.2 (85.5) & 62.4 (62.5) & 72.8 (83.6) & 73.8 (70.9) & 85.7 (80.7) & \textbf{88.3} (\textbf{83.7}) & \textbf{92.4} (\textbf{85.6}) & \textbf{90.1} (\textbf{80.4}) & \underline{92.8} (84.5) \\ 
LCUBE & 55.7 (52.5) & 31.8 (31.3) & 43.5 (33.1) & 87.8 (\textbf{82.1}) & \textbf{100.0} (\textbf{100.0}) & \textbf{100.0} (\textbf{100.0}) & \textbf{100.0} (\textbf{100.0}) & \textbf{100.0} (\textbf{100.0}) & \textbf{100.0} (\textbf{100.0}) & \underline{71.4} (62.3) & 78.1 (66.7) & \underline{88.7} (\underline{76.8}) & \textbf{95.4} (\underline{86.0}) \\
SLOPE & 59.3 (57.0) & \underline{73.3} (\underline{67.3}) & \underline{96.9} (88.7) & 67.1 (62.3) & 9.2 (18.0) & 23.4 (28.0)& 12.6 (21.0) & 11.0 (17.0) & 1.1 (7.0) & 47.9 (45.0) & 57.2 (54.0) & 45.1 (46.0) & 44.5 (47.0) \\
\bottomrule
\end{tabular}
}
\end{table}

Additionally, it is also important to note that the four RDMDL versions achieve comparable performance across both the T\"ubingen and most synthetic benchmarks, suggesting its performance is not overly sensitive to the choice of distortion.

\begin{figure}[h]
    \centering
    \includegraphics[width=0.45\linewidth]{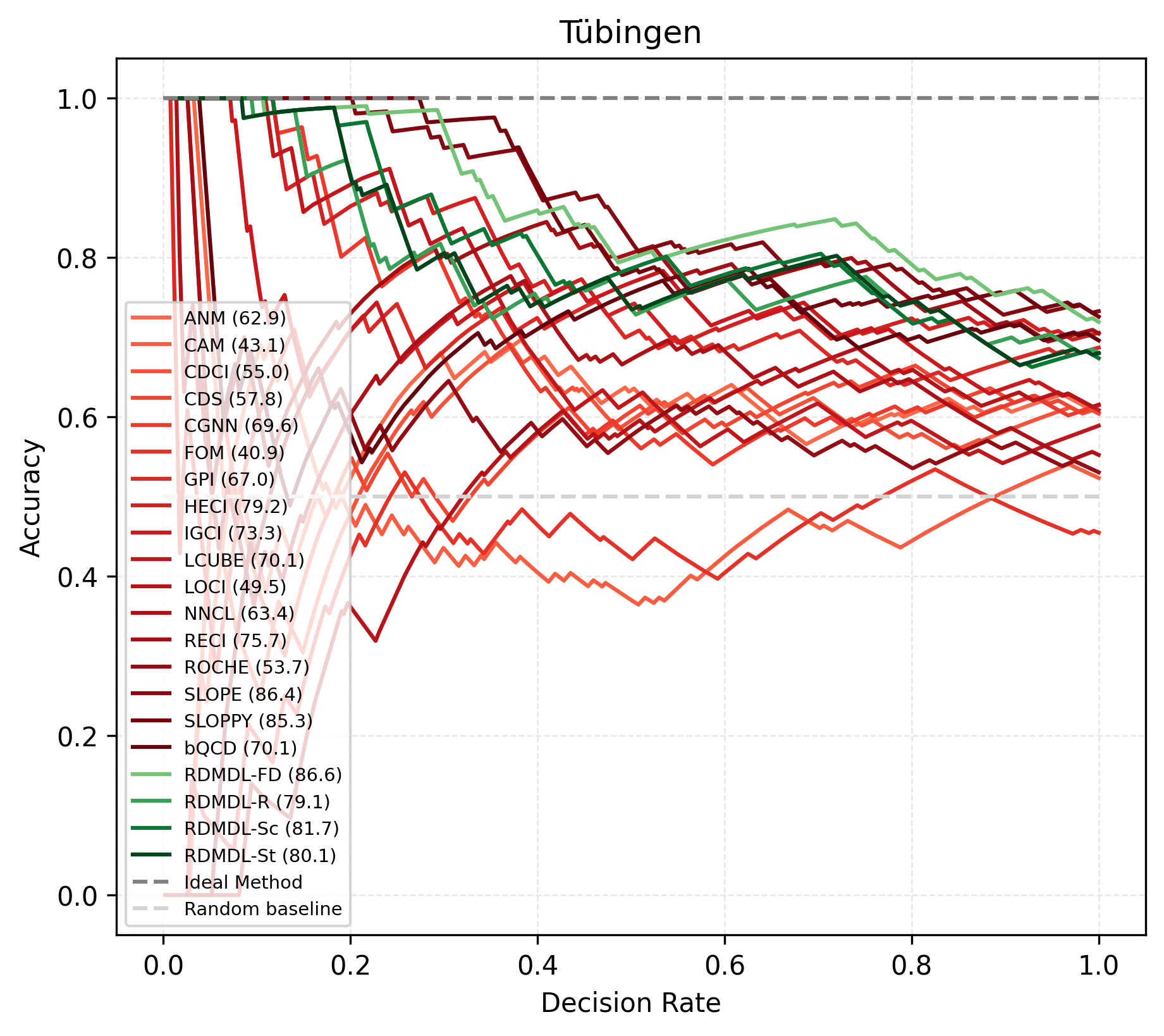}
    \includegraphics[width=0.45\linewidth]{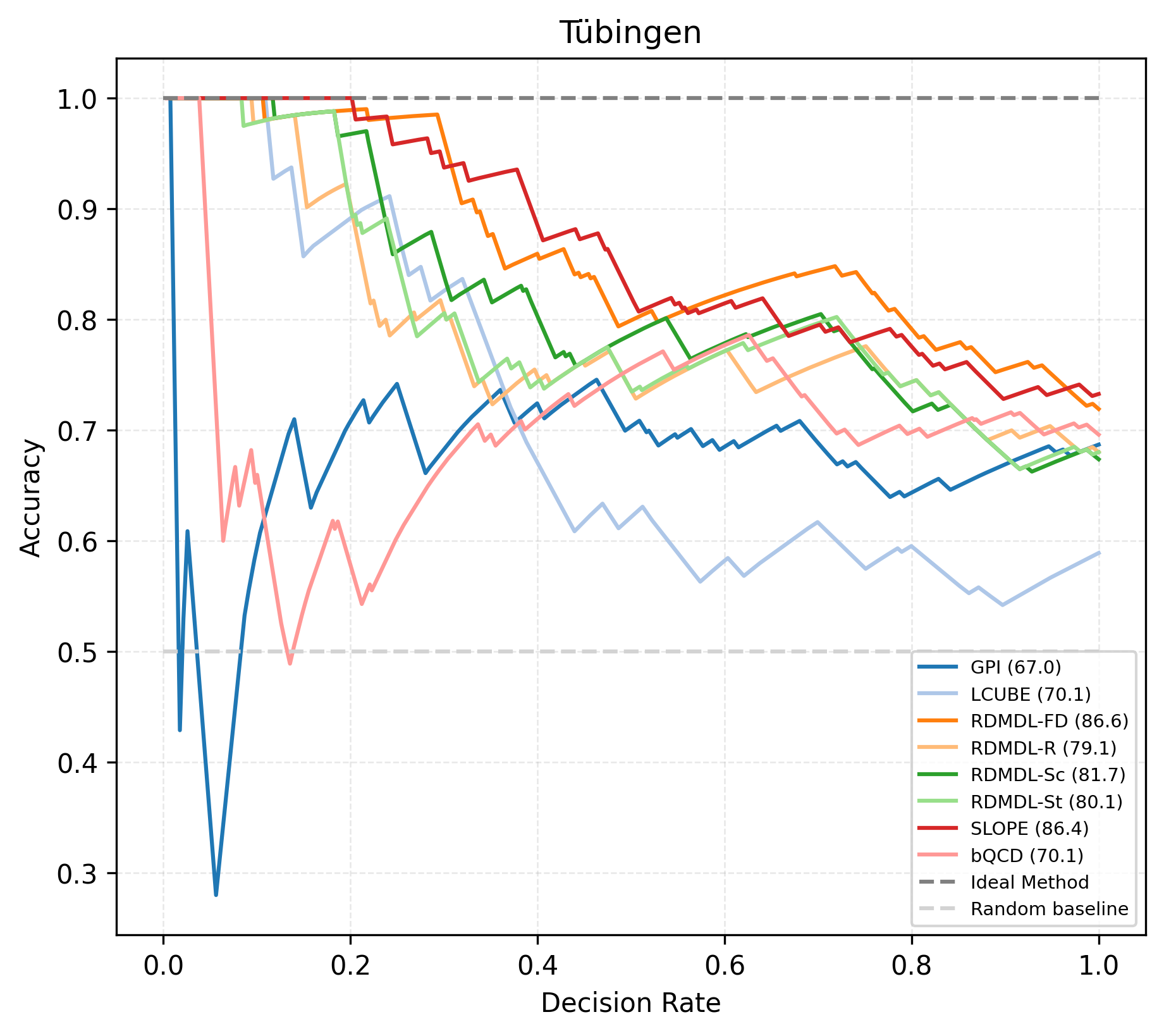}
    \caption{AUDRC curves in the T\"ubingen benchmark. Left: RDMDL (green) compared to all other implemented methods (shades of red); right: analysed MDL causal discovery methods.
    }
    \label{rdmdlcurves}
\end{figure}

Figure \ref{rdmdlcurves} displays the AUDRC for the four versions of RDMDL, compared to other methods on the T\"ubingen benchmark. In it, it is clear that RDMDL performs consistently above average across all decision rates. This is an even stronger result than the values of AUROC, accuracy, and AUDRC shown in Table \ref{lisbonandtuebingenres}, showing that, regardless of the decision rate, RDMDL is a leading method for causal discovery.  

Furthermore, given that one of key contributions of this work is to propose an MDL method for bivariate causal discovery that performs an adequate estimate of $L(X)$, we now analyze the relevance of $L(X)$ in the overall performance of RDMDL. 
Table \ref{colorstable1} studies how informative $L(X)$ and $L(Y | X)$ are individually, and how they interact to yield the result on the T\"ubingen benchmark. Understanding this behaviour is crucial, as it reflects the principle that MDL seeks to exploit (see Equation \eqref{eq:difs}). Each entry in both tables contains two values: on the left, the percentage of weighted guesses where the difference $(L(X)-L(Y))$ dominates the decision (higher absolute value), while on the right are those where $(L(Y | X) - L(X|Y))$ dominates. 

The first clear conclusion is that both differences are individually good predictors. The second one is that there are a few differences between the four versions \footnote{Note that the computation using Sturges and Rice methods yields the same results. This is not, however, necessarily always true; their difference is not strong enough to meaningfully cause any difference.} of RDMDL. The third, and maybe most crucial, is that both differences are responsible for a significant part of the decisions, thus both contribute actively and positively to the predictions. 

Lastly, although each individual predictor performs well and contributes equally to the output, they do not appear to reinforce one another. Recall that a central principle of MDL is that a joint distribution producing a negative prediction in one estimator should correspondingly generate an even stronger positive prediction in the other. However, this effect is not observed in practice. This would manifest as the correct predictor outnumbering the wrong one in the off-diagonal entries.

\vspace{-0.5cm}

\begin{table}[h]
\centering
\caption{Confusion matrix: in each cell, on the left, the result when taken by $(L(X)-L(Y))$ and on the right when $(L(Y | X)-L(X|Y))$ dominates. Green indicates a correct decision and red an incorrect decision. Top-left table: RDMDL-FD; top-right table: RDMDL-R; bottom-left table: RDMDL-Sc; bottom-right table: RDMDL-St.}
\label{colorstable1}
\resizebox{0.49\textwidth}{!}{
\begin{tabular}{|c|c|c|c|c|} 
\hline
\multicolumn{2}{|c|}{\multirow{2}{*}{\textbf{Predictors-FD}}} & \multicolumn{2}{c|}{\textbf{$L(Y | X)$}} & \multirow{2}{*}{\textbf{Marginals}} \\
\cline{3-4}
\multicolumn{2}{|c|}{} & \textbf{correct} & \textbf{incorrect} & \\
\hline
\multirow{2}{*}{\textbf{$L(X)$}} & \textbf{correct} & \textcolor{green}{38.2} / \textcolor{green}{16.2} & \textcolor{green}{15.8} / \textcolor{red}{4.8} & \textcolor{green}{70.2} / \textcolor{red}{4.8} \\
\cline{2-5}
& \textbf{incorrect} & \textcolor{red}{8.1} / \textcolor{green}{1.7} & \textcolor{red}{8.3} / \textcolor{red}{6.9} & \textcolor{green}{1.7} / \textcolor{red}{23.3} \\
\hline
\multicolumn{2}{|c|}{\textbf{Marginals}} & \textcolor{green}{56.1} / \textcolor{red}{8.1} & \textcolor{green}{15.8} / \textcolor{red}{20.0} & \\
\hline
\end{tabular}}
\resizebox{0.49\textwidth}{!}{
\begin{tabular}{|c|c|c|c|c|} 
\hline
\multicolumn{2}{|c|}{\multirow{2}{*}{\textbf{Predictors-R}}} & \multicolumn{2}{c|}{\textbf{$L(Y | X)$}} & \multirow{2}{*}{\textbf{Marginals}} \\
\cline{3-4}
\multicolumn{2}{|c|}{} & \textbf{correct} & \textbf{incorrect} & \\
\hline
\multirow{2}{*}{\textbf{$L(X)$}} & \textbf{correct} & \textcolor{green}{23.3} / \textcolor{green}{21.5} & \textcolor{green}{11.3} / \textcolor{red}{7.6} & \textcolor{green}{56.1} / \textcolor{red}{7.6} \\
\cline{2-5}
& \textbf{incorrect} & \textcolor{red}{7.4} / \textcolor{green}{12.0} & \textcolor{red}{4.8} / \textcolor{red}{12.2} & \textcolor{green}{12.0} / \textcolor{red}{24.4} \\
\hline
\multicolumn{2}{|c|}{\textbf{Marginals}} & \textcolor{green}{56.8} / \textcolor{red}{7.4} & \textcolor{green}{11.3} / \textcolor{red}{24.6} & \\
\hline
\end{tabular}}

\vspace{0.1cm}
\resizebox{0.49\textwidth}{!}{ 
\begin{tabular}{|c|c|c|c|c|}
\hline
\multicolumn{2}{|c|}{\multirow{2}{*}{\textbf{Predictors-Sc}}} & \multicolumn{2}{c|}{\textbf{$L(Y | X)$}} & \multirow{2}{*}{\textbf{Marginals}} \\
\cline{3-4}
\multicolumn{2}{|c|}{} & \textbf{correct} & \textbf{incorrect} & \\
\hline
\multirow{2}{*}{\textbf{$L(X)$}} & \textbf{correct} & \textcolor{green}{27.1} / \textcolor{green}{26.4} & \textcolor{green}{10.6} / \textcolor{red}{4.8} & \textcolor{green}{64.1} / \textcolor{red}{4.8} \\
\cline{2-5}
& \textbf{incorrect} & \textcolor{red}{7.4} / \textcolor{green}{3.2} & \textcolor{red}{10.0} / \textcolor{red}{10.5} & \textcolor{green}{7.4} / \textcolor{red}{27.9} \\
\hline
\multicolumn{2}{|c|}{\textbf{Marginals}} & \textcolor{green}{56.7} / \textcolor{red}{7.4} & \textcolor{green}{10.6} / \textcolor{red}{25.3} & \\
\hline
\end{tabular}}
\resizebox{0.49\textwidth}{!}{
\vspace{0.4cm}
\begin{tabular}{|c|c|c|c|c|}
\hline
\multicolumn{2}{|c|}{\multirow{2}{*}{\textbf{Predictors-St}}} & \multicolumn{2}{c|}{\textbf{$L(Y | X)$}} & \multirow{2}{*}{\textbf{Marginals}} \\
\cline{3-4}
\multicolumn{2}{|c|}{} & \textbf{correct} & \textbf{incorrect} & \\
\hline
\multirow{2}{*}{\textbf{$L(X)$}} & \textbf{correct} & \textcolor{green}{23.3} / \textcolor{green}{21.5} & \textcolor{green}{11.3} / \textcolor{red}{7.6} & \textcolor{green}{56.1} / \textcolor{red}{7.6} \\
\cline{2-5}
& \textbf{incorrect} & \textcolor{red}{7.4} / \textcolor{green}{12.0} & \textcolor{red}{4.8} / \textcolor{red}{12.2} & \textcolor{green}{12.0} / \textcolor{red}{24.4} \\
\hline
\multicolumn{2}{|c|}{\textbf{Marginals}} & \textcolor{green}{56.8} / \textcolor{red}{7.4} & \textcolor{green}{11.3} / \textcolor{red}{24.6} & \\
\hline
\end{tabular}}
\end{table}

\section{Conclusions and Future Work}
This work revisited the fundamental rationale behind MDL for bivariate causal discovery. A detailed analysis of existing methods showed that the estimation of $L(X)$, the description length of the cause variable, has been largely overlooked, with decisions often effectively relying only on the complexity of the causal mechanism. Motivated by this gap, we proposed a novel approach to estimate $L(X)$ via rate-distortion theory, combined with a traditional MDL-based approach to computing $L(Y | X)$. The computation of $L(X)$ is based on an informative choice of distortion, stemming from histogram-based density estimation rules, and an approximation of the rate-distortion dimension under the asymptotic assumptions.
This yields our new method, rate-distortion MDL (RDMDL), which achieves competitive performance on the Tübingen and most synthetic datasets.

A major avenue for future work is to  extend the rate-distortion estimation, or a similar theoretically grounded lower bound on Kolmogorov complexity, to measure 
$L(Y | X)$. This would not only provide a more coherent integration with the estimation of $L(X)$, but also hopefully better performance on real data.  Additionally, the deduction of specific scenarios where RDMDL is guaranteed to be correct would contribute to reinforce the validity of the current approach.

\acks{Work partially funded by \textit{Fundação para a Ciência e a Tecnologia} (Portugal), project \url{doi.org/10.54499/UID/50008/2025}. We also acknowledge support from \textit{Feedzai Lda} (Portugal).}

\bibliography{sample}

\appendix
\clearpage
\section{Full Algorithm} \label{algorithm_appendix}

This appendix provides the complete description of our implementation. Algorithm \ref{rdmdl_alg} outlines the RDMDL procedure. The initial step involves scaling both the input data $X$ and $Y$, using either min-max scaling or normalization (subtracting the mean and dividing by the standard deviation). Subsequently, the distortion parameter $D$ is selected according to the criteria detailed in Subsection \ref{possible distortions}. The algorithm then computes the respective Minimum Description Length (MDL) encoding lengths for both directions: $L(X)$, $L(Y \mid X)$, $L(Y)$, and $L(X \mid Y)$, utilizing Algorithms \ref{lx_alg} and \ref{lyx_alg}. Finally, it calculates the total encoding length for both directions, and outputs their difference, which is then divided by $N$. This normalization serves to express the score as the average difference in the number of bits per observation pair, providing a better interpretation and facilitating the use of rank-based metrics (such as AUROC and AUDRC) for assessing the confidence level of RDMDL. Consequently, the output score represents the average difference in the number of bits per pair required to encode the data under the causal assumption in each direction.

\begin{algorithm}[H]
\LinesNumbered
\caption{RDMDL}
\label{rdmdl_alg}
\KwIn{$X$, $Y$}
\KwOut{$S_{X,Y}$}

Scale $X$ and $Y$ using min-max scaling or normalization.

Choose the distortion parameter $D$. 

Compute the total encoding lengths for the models in each direction using Algorithms \ref{lx_alg} and \ref{lyx_alg}:
$$L_{X \to Y} = L_X(X) + L_{Y\mid X}(X,Y)$$
$$L_{Y \to X} = L_X(Y) + L_{Y\mid X}(Y,X).$$

Compute the final score as the average encoding length per data point:
$$S_{X,Y} \leftarrow  \frac{L_{Y \to X}-L_{X \to Y}}{N_X}.$$

\end{algorithm}
\noindent 

\vspace{-1cm}

\subsection{$L(X)$ Algorithm}

This subsection details the computation of the encoding length $L(X)$. In general, the procedure first calculates the information dimension $\dim_X$, and then uses it to compute the total encoding length based on Equation \eqref{lx_calc}. The function used to compute the information dimension is described in Algorithm \ref{informationdimensionalg}, while the entropy calculation via binning is detailed in Algorithm \ref{entropyalg}.

\begin{algorithm}[H]
\LinesNumbered
\caption{$L_X$ (from RDMDL)}
\label{lx_alg}
\KwIn{$X$, $D$}
\KwOut{$L_X$}

Compute the Information Dimension using Algorithm \ref{informationdimensionalg}:
\[
\dim_X \leftarrow \dim(X).
\]

Compute the total encoding length for the data:
$$L_X \leftarrow N_X \, \frac{\dim_X}{2}\,\log\left(\frac{1}{D}\right).$$

\end{algorithm}
\noindent In Algorithm \ref{lx_alg}, $D$ is the chosen distortion parameter and $N_X$ is the number of observations in $X$.

\begin{algorithm}[H]
\LinesNumbered
\caption{Information Dimension $\dim(X)$ (from RDMDL)}
\label{informationdimensionalg}
\KwIn{$X$, $H_X(\cdot)$, $\epsilon_{\min}$, $\epsilon_{\max}$, and $N_\epsilon$}
\KwOut{$\mathrm{dim}_X$}

$E$ = $\text{logspace}(\epsilon_{\min}, \epsilon_{\max}, N_\epsilon)$.

$H_i \leftarrow H_X(\epsilon_i), \forall \, \epsilon_i \in E$. \Comment{In accordance with Algorithm \ref{entropyalg}.}

Fit a linear model to the points $(\log \epsilon_i, H_i)$: \Comment{Using Numpy's polyfit.}
\[
H_i \approx a\,\log \epsilon_i + b
\]
with least squares regression.

Compute the information dimension: 
\[
\dim_X \leftarrow -a.
\]
\end{algorithm}
\noindent In Algorithm \ref{informationdimensionalg}, the function logspace stands for the standard Numpy function.

\begin{algorithm}[H]
\LinesNumbered
\caption{Entropy $H$ (from RDMDL)}
\label{entropyalg}
\KwIn{$X$, $\epsilon_i$}
\KwOut{$H_{X,\epsilon_i}$}

$b \leftarrow \text{linspace} (\min(X),\max(X), \frac{1}{\epsilon_i} +1 )$ \Comment{Define bin edges}

$c \leftarrow \text{histogram} (X, b)$ 

$p \leftarrow \frac{c_{c>0}}{N_X}$ \Comment{Compute non-zero probabilities}

$H_{X,\epsilon_i} \leftarrow -\sum p_i \cdot \log (p_i)$ 

\end{algorithm}
\noindent In Algorithm \ref{entropyalg}, both linspace and histogram refer to the standard Numpy functions.

\subsection{$L(Y\mid X)$ Algorithm}

The general procedure for computing the conditional encoding length $L(Y|X)$ involves fitting three different sets of functional models (polynomial, inversely proportional, and exponential/logarithmic), and then selecting the model that yields the shortest description length. For each model set, the process simultaneously determines the best least squares fit, by minimizing the total encoding length for both the residuals and the model parameters.

\begin{algorithm}[H]
\LinesNumbered
\caption{$L_{Y \mid X}$ (from RDMDL)}
\label{lyx_alg}
\KwIn{$X$, $Y$}
\KwOut{$L_{Y\mid X}$}

$L_p = \text{fit}_{\text{polynomial}}(X,Y)$ \Comment{Check Algorithm \ref{fitalg}}

$L_i = \text{fit}_{\text{inversely}}(X,Y)$

$L_e = \text{fit}_{\text{explog}}(X,Y)$

Return the encoding length of the best-fitting model:
$$L_{Y\mid X} \leftarrow \min\left(L_p , L_i,L_e\right)$$

\end{algorithm}

\vspace{-0.3cm}

\noindent  In Algorithm \ref{lyx_alg}, $L_p$, $L_i$, and $L_e$ stand for the minimum obtained length when encoding the causal mechanism with polynomial, inversely proportional, and exponential/logarithmic functions respectively. The three different lengths are computed in accordance with Algorithm \ref{fitalg}, by passing a different sets of possible families of functions in the argument ($F$).

\begin{algorithm}[H]
\LinesNumbered
\caption{$\text{fit}$ (from RDMDL)}
\label{fitalg}
\KwIn{$X$, $Y$, $F = \{f_1, \ldots , f_m\}$}
\KwOut{$L$}

For each function $f$ in the set of functions $F$, compute:
$$r_f , \theta_f \leftarrow \text{LS}_f (X,Y),$$
where $\text{LS}_f$ denotes the best least squares fit for the function $f$, yielding the residuals $r_f$ and model parameters $\theta_f$.

Then, for each function, compute the encoding length of the residuals using a Gaussian distribution assumption:
$$L(r_f) \leftarrow \frac{N_r}{2}\, \log \left(2 \pi \frac{\sum r_f^2 }{N_r}\right) + \frac{N_r}{2\, \log(2)}.$$

Compute the encoding length for the model parameters:
$$L(\theta_f) \leftarrow \frac{N_{\theta_f}}{2} \, \log (N_r).$$

Lastly, the overall length $L$ is the minimum of the sum of the residual and model parameter lengths obtained by all tested functions:
$$L \leftarrow \min_f (L(r_f) + L(\theta_f)).$$
\end{algorithm}

\vspace{-0.3cm}

\noindent In Algorithm \ref{fitalg}, $N_r$ is the length of the residual vector ($N_r=N_X=N_Y$) and $N_{\theta_f}$ is the number of parameters in model $f$.

\newpage
\section{Additional results}

\subsection{Unimplemented results}\label{unimplementedres}

In this Appendix, the results of 9 different unimplemented causal discovery methods are reported: COMIC \citep{bayesianmdl}, EMD \citep{rkhs}, GPLVM \citep{bayes}, GRCI \citep{grci}, LiNGAM \citep{shimizu2006linear}, PNL \citep{zhang2010distinguishing}, CURE \citep{cure}, and GRAN \citep{gran}. These methods were not implemented since either 1) no publicly available implementation was found or 2) the method required training. Nevertheless, the reported results of each method (alongside its sources) are reported in Tables \ref{tab:accuracy_comparison}, \ref{tab:AUROC_comparison}, and \ref{tab:AUDRC_comparison}. There are results reported for the real-world T\"ubingen dataset \citep{tuebingen} and the synthetic datasets of \citet{tuebingenresults} (CE), \citet{tuebingen} (SIM), and \citet{bcqd} (ANLSMN).

\begin{table}[H]
\centering
\caption{Accuracy (in \%) obtained by the different unimplemented methods for all evaluated datasets.}
\label{tab:accuracy_comparison}
\resizebox{\linewidth}{!}{%
\small
\begin{tabular}{l|l|ccccccccccccccc}
\toprule
Method & Source & AN & AN-s & CE-Cha & CE-Multi & CE-Net & LS & LS-s & MN-U & Multi & Net & SIM & SIM-G & SIM-c & SIM-ln & T\"ubingen \\ \hline
\midrule
COMIC  &         COMIC &  100 &  100 &             43 &  79 &  78 &  100 &  100 &  100 &                   &                   &              57 &              78 &              54 &             89 &              67 \\
EMD    &          bQCD &           36 &           33 &                  &                   &                   &            60 &           42 &           83 &                   &                   &              45 &              58 &               40 &             52 &              68 \\
GPLVM  &  GPLVM, COMIC &  100 &  100 &                  &                   &                   &  100 &  100 &  100 &                   &                   &  83 &              92 &              79 &  90 &                   \\
GRCI   &          LOCI &  100 &           94 &  70 &                   &                   &           98 &           87 &           88 &  77 &  85 &              77 &               70 &              77 &             77 &  82 \\
LiNGAM &          bQCD &              0 &           4 &                  &                   &                   &           7 &           3 &              0 &                   &                   &              42 &              25 &              53 &             31 &              42 \\
PNL    &          bQCD &           96 &           63 &                  &                   &                   &           91 &           44 &           66 &                   &                   &               70 &              64 &              65 &             61 &              73 \\
CURE   &          bQCD &                &                &                  &                   &                   &                &                &                &                   &                   &                   &                   &                   &                  &              61 \\
GRAN   &          bQCD &           5 &           6 &                  &                   &                   &           11 &            20 &            50 &                   &                   &              48 &              37 &              44 &             43 &               50 \\
\bottomrule
\end{tabular}}
\normalsize
\end{table}

\begin{table}[H]
\centering
\caption{AUROC (in \%) obtained by the different unimplemented methods for all evaluated datasets.}
\label{tab:AUROC_comparison}
\resizebox{\linewidth}{!}{%
\small
\begin{tabular}{l|l|cccccccccccccc}
\toprule
Method & Source & AN & AN-s & CE-Cha & CE-Gauss & CE-Multi & CE-Net & LS & LS-s & MN-U & SIM & SIM-G & SIM-c & SIM-ln & T\"ubingen \\ \hline
\midrule
COMIC  &         COMIC &  100 &  100 &               47 &                    &               91 &               84 &  100 &  100 &  100 &              58 &              85 &              59 &              95 &               75 \\
EMD    &          bQCD &                &                &                    &                    &                    &                    &                &                &                &                   &                   &                   &                   &               43 \\
GPLVM  &  GPLVM, COMIC & 100 &  100 &  82 &  89 &  98 &  99 &  100 &  100 &  100 &  92 &  99 &  91 &  97 &  78 \\
LiNGAM &          bQCD &                &                &                    &                    &                    &                    &                &                &                &                   &                   &                   &                   &                30 \\
PNL    &          bQCD &                &                &                    &                    &                    &                    &                &                &                &                   &                   &                   &                   &                70 \\
CURE   &          bQCD &                &                &                    &                    &                    &                    &                &                &                &                   &                   &                   &                   &               64 \\
GRAN   &          bQCD &                &                &                    &                    &                    &                    &                &                &                &                   &                   &                   &                   &               47 \\
\bottomrule
\end{tabular}}
\normalsize
\end{table}

\begin{table}[H]
\centering
\caption{AUDRC (in \%) obtained by the different unimplemented methods for all evaluated datasets.}
\label{tab:AUDRC_comparison}
\resizebox{\linewidth}{!}{%
\small
\begin{tabular}{l|l|ccccccccccccc}
\toprule
Method & Source & AN & AN-s & CE-Cha & LS & LS-s & MN-U & Multi & Net & SIM & SIM-G & SIM-c & SIM-ln & T\"ubingen \\ \hline
\midrule
GRCI &  LOCI &  100 &  100 &  71 &  100 &  95 &  97 &  74 &  96 &  90 &  88 &  92 &  92 &  73 \\
\bottomrule
\end{tabular}}
\normalsize
\end{table}

\newpage

\subsection{Synthetic results}
\label{syntheticappendix}

In this Appendix, the evaluation of the seventeen implemented bivariate causal discovery methods, alongside the four implementations of RDMDL, is reported for the following three sets of synthetic cause-effect benchmarks: the CE datasets of \citet{tuebingenresults}, the SIM datasets of \citet{tuebingen}, and the ANLSMN datasets of \citet{bcqd}.

\begin{table}[h]
\centering
\caption{Performance of the sixteen implemented causal discovery methods, alongside the four versions of RDMDL, in three sets of synthetic benchmarks, evaluated using AUROC (accuracy). The value in bold corresponds to the best result in each column, and those underlined to the second best.}
\label{tab:synthetic_performance}
\resizebox{\textwidth}{!}{%
\begin{tabular}{lccccccccccccc}
\toprule
Method & CE-Cha & CE-Gauss & CE-Multi & CE-Net & AN & AN-s & LS & LS-s & MN-U & SIM & SIM-c & SIM-G & SIM-ln \\
\midrule
RDMDL-FD & 62.7 (60.7) & 64.7 (58.7) & \textbf{97.0 }(\underline{89.3}) & 75.2 (68.3) & 51.2 (49.0) & 43.3 (45.0) & 49.0 (47.0) & 13.4 (17.0) & 0.9 (6.0) & 53.6 (47.0) & 59.6 (57.0) & 59.4 (55.0) & 87.2 (77.0) \\
RDMDL-R & 64.2 (60.3) & 64.5 (59.0) & 96.5 (88.3) & 81.1 (72.0) & 84.5 (74.0) & 52.3 (51.0) & 77.0 (68.0) & 6.9 (13.0) & 1.3 (7.0) & 61.9 (56.0) & 66.9 (65.0) & 66.2 (57.0) & 91.4 (79.0) \\
RDMDL-Sc & 63.7 (59.3) & 69.4 (62.7) & \underline{96.9} (89.0) & 77.5 (69.3) & 58.6 (52.0) & 34.7 (34.0) & 53.5 (51.0) & 4.8 (12.0) & 0.8 (4.0) & 58.1 (53.0) & 63.9 (62.0) & 60.1 (54.0) & 88.6 (78.0) \\
RDMDL-St & 64.0 (60.3) & 65.1 (59.0) & 96.5 (88.3) & 81.7 (72.3) & 86.8 (76.0) & 53.7 (51.0) & 78.5 (69.0) & 5.8 (13.0) & 1.4 (7.0) & 62.8 (56.0) & 67.6 (66.0) & 66.4 (58.0) & 91.6 (79.0) \\
\hline
ANM & 34.3 (40.0) & 43.3 (45.3) & 48.7 (51.0) & 56.3 (59.0) & 52.8 (51.2) & 52.9 (55.8) & 59.4 (57.8) & 52.3 (57.8) & 63.5 (62.2) & 49.1 (46.3) & 47.8 (43.6) & 27.5 (37.2) & 60.7 (56.2) \\
bQCD & 56.3 (55.0) & 48.1 (54.3) & 53.2 (49.0) & \underline{90.7} (\underline{81.3}) & \textbf{100.0} (\textbf{100.0}) & 90.5 (83.0) & \textbf{100.0} (\textbf{100.0}) & \textbf{100.0} (\textbf{100.0}) & \textbf{100.0} (\textbf{100.0}) & 69.9 (68.0) & 80.8 (\underline{77.0}) & 71.7 (67.0) & 91.9 (\textbf{87.0}) \\
CAM & 58.6 (53.3) & 30.0 (25.0) & 38.3 (35.3) & 81.6 (78.3) & \textbf{100.0}(\textbf{100.0})& \textbf{100.0}(\textbf{100.0})& 99.9 (98.0) & 36.9 (52.0) & 81.2 (86.0) & 62.3 (56.0) & 64.3 (62.0) & 88.2 (\underline{82.0}) & 86.0 (\textbf{87.0}) \\
CDCI & \textbf{72.2} (\underline{66.7}) & \textbf{91.7} (83.0) & 95.7 (90.0) & \textbf{91.8} (80.7) & \textbf{100.0}(\textbf{100.0})& 99.1 (95.0) & \textbf{100.0}(\textbf{100.0})& 98.9 (95.0) & 99.5 (95.0) & \underline{87.2} (\underline{81.0}) & 84.1 (72.0) & 81.2 (73.0) & 85.0 (76.0) \\
CDS & \underline{69.2} (\textbf{71.0}) & \underline{90.6} (\textbf{84.0}) & 41.2 (43.7) & 88.6 (78.3) & \textbf{100.0}(99.0) & 99.9 (99.0) & 84.7 (76.0) & 1.3 (5.0) & 71.7 (70.0) & 82.1 (71.0) & \underline{84.7} (76.0) & 79.7 (73.0) & 78.5 (65.0) \\
CGNN & 55.5 (56.0) & 37.4 (42.3) & 80.1 (68.0) & 60.0 (56.3) & 77.9 (67.0) & 89.8 (81.0)& 78.8 (70.0) & 86.6 (75.0) & 2.7 (8.0) & 42.0 (42.0) & 51.3 (53.0) & 84.9 (73.0) & 77.7 (68.0) \\
FOM & 42.0 (44.3) & 56.1 (53.0) & 10.4 (20.0) & 27.0 (35.0) & 94.9 (86.0) & \textbf{100.0}(\textbf{100.0})& 92.9 (93.0) & 99.3 (99.0) & \textbf{100.0}(\textbf{100.0})& 51.1 (52.0) & 46.2 (45.0) & 25.1 (33.0) & 36.9 (43.0) \\
GPI & 64.9 (62.4) & 89.1 (\underline{83.2}) & 72.2 (71.5) & 88.8 (81.3) & 79.2 (85.5) & 62.4 (62.5) & 72.8 (83.6) & 73.8 (70.9) & 85.7 (80.7) & \textbf{88.3} (\textbf{83.7}) & \textbf{92.4} (\textbf{85.6}) & \underline{90.1} (80.4) & \underline{92.8} (84.5) \\ 
HECI & 55.6 (56.7) & 51.9 (48.3) & 97.5 (90.7) & 79.4 (72.3) & 99.9 (98.0) & 58.4 (55.0) & 99.1 (92.0) & 62.4 (55.0) & 22.9 (33.0) & 53.8 (49.0) & 59.4 (55.0) & 63.5 (56.0) & 78.9 (65.0) \\
IGCI & 55.9 (55.0) & 16.0 (21.3) & 78.0 (68.0) & 57.2 (57.0) & 98.5 (89.0) & 99.5 (97.0)& 98.6 (95.0) & 98.7 (94.0) & 92.9 (86.0) & 32.2 (36.0) & 37.8 (42.0) & \textbf{94.4} (\textbf{86.0}) & 65.3 (59.0) \\
LCUBE & 55.7 (52.5) & 31.8 (31.3) & 43.5 (33.1) & 87.8 (\textbf{82.1}) & \textbf{100.0} (\textbf{100.0}) & \textbf{100.0} (\textbf{100.0}) & \textbf{100.0} (\textbf{100.0}) & \textbf{100.0} (\textbf{100.0}) & \textbf{100.0} (\textbf{100.0}) & 71.4 (62.3) & 78.1 (66.7) & 88.7 (76.8) & \textbf{95.4} (86.0) \\
LOCI & 56.9 (57.3) & 82.7 (73.0) & 71.4 (70.0) & 89.9 (77.7) & \textbf{100.0}(\textbf{100.0})& 99.9 (99.0) & 91.1 (85.0) & 58.4 (58.0) & 99.7 (96.0) & 82.7 (72.0) & 79.6 (68.0) & 81.9 (73.0) & 81.3 (74.0) \\
NNCL & 49.6 (50.0) & 48.5 (48.7) & 57.6 (55.7) & 79.9 (66.7) & 93.4 (84.0) & 26.8 (33.0) & 86.2 (84.0) & 17.6 (19.0) & 78.7 (60.0) & 67.1 (64.0) & 68.5 (59.0) & 71.7 (62.0) & 69.6 (54.0) \\
RECI & 58.8 (56.0) & 71.1 (64.3) & 94.8 (85.3) & 65.8 (60.3) & 8.4 (18.0) & 24.8 (35.0) & 15.2 (22.0) & 39.0 (44.0) & 5.6 (13.0) & 45.2 (44.0) & 55.7 (53.0) & 43.7 (39.0) & 38.1 (44.0) \\
ROCHE & 54.7 (51.3) & 21.7 (28.3) & 86.2 (73.7) & 73.9 (69.0) & \textbf{100.0}(\textbf{100.0})& \textbf{100.0}(\textbf{100.0})& \textbf{100.0}(\textbf{100.0})& 99.0 (99.0) & \textbf{100.0} (\textbf{100.0}) & 43.0 (45.0) & 47.6 (46.0) & 83.7 (72.0) & 81.7 (71.0) \\
SLOPE & 59.3 (57.0) & 73.3 (67.3) & \underline{96.9} (88.7) & 67.1 (62.3) & 9.2 (18.0) & 23.4 (28.0)& 12.6 (21.0) & 11.0 (17.0) & 1.1 (7.0) & 47.9 (45.0) & 57.2 (54.0) & 45.1 (46.0) & 44.5 (47.0) \\
SLOPPY & 58.5 (54.0) & 68.3 (59.0) & 96.6 (\textbf{90.3}) & 70.1 (62.3) & 11.2 (17.0) & 15.4 (20.0) & 14.3 (22.0) & 6.6 (10.0) & 0.5 (5.0) & 47.7 (46.0) & 56.9 (53.0) & 46.3 (46.0) & 56.9 (54.0) \\
\bottomrule
\end{tabular}
}
\end{table}

\end{document}